\documentclass[runningheads]{llncs}

\makeatletter
\newcommand{\@chapapp}{\relax}%
\makeatother

\usepackage{times}
\usepackage{latexsym}
\usepackage{url}
\usepackage{soul}
\usepackage{graphicx}
\usepackage[toc,page]{appendix}
\usepackage{bbm}
\usepackage{booktabs}
\usepackage{amsmath,amsfonts,amssymb,amscd}
\usepackage{float}
\usepackage{soul}
\usepackage{multirow}

\usepackage{subfig}
\usepackage{xcolor}
\usepackage{hyperref}
\usepackage[nameinlink]{cleveref}
\usepackage[ruled,vlined]{algorithm2e}
\SetKwComment{Comment}{$\triangleright$\ }{}
\usepackage{adjustbox}
\usepackage{rotating}
\usepackage[numbers]{natbib}
\usepackage{array}
\usepackage{placeins}

\newcommand{\comment}[1]{} 

\title{Bringing a Ruler Into the Black Box: Uncovering Feature Impact from Individual Conditional Expectation Plots\texorpdfstring{\thanks{We thank David Rosenberg for his insightful feedback, engagement, and advice in shaping this project from proposal to paper as well for introducing us to the original ICE paper. We thank Anu-Ujin Gerelt-Od without whom this project would not be possible. Last but not least, we thank Lee Kho for her valuable input, ideas, and support.}}{*}}

\author{Andrew Yeh\inst{1,2} \and
Anhthy Ngo\inst{1,3}}
\institute{New York University, New York NY 10011, USA \email{\{ay1626,an3056\}@nyu.edu}
\and The Wharton School of the University of Pennsylvania, Philadelphia PA 19104, USA \email{ayeh21@upenn.edu}
\and The MITRE Corporation, McLean VA 22102, USA \email{ango@mitre.org}}

\begin{document}

\maketitle

\begin{abstract}
As machine learning systems become more ubiquitous, methods for understanding and interpreting these models become increasingly important. In particular, practitioners are often interested both in what features the model relies on and how the model relies on them -- the feature's impact on model predictions. Prior work on feature impact including partial dependence plots (PDPs) and Individual Conditional Expectation (ICE) plots has focused on a visual interpretation of feature impact. We propose a natural extension to ICE plots with ICE feature impact, a model-agnostic, performance-agnostic feature impact metric drawn out from ICE plots that can be interpreted as a close analogy to linear regression coefficients. Additionally, we introduce an in-distribution variant of ICE feature impact to vary the influence of out-of-distribution points as well as heterogeneity and non-linearity measures to characterize feature impact. Lastly, we demonstrate ICE feature impact's utility in several tasks using real-world data.
\end{abstract}

\section{Introduction}
\label{sec:intro}

\begin{figure*}[t]
    \centering
    \includegraphics[scale = 0.9]{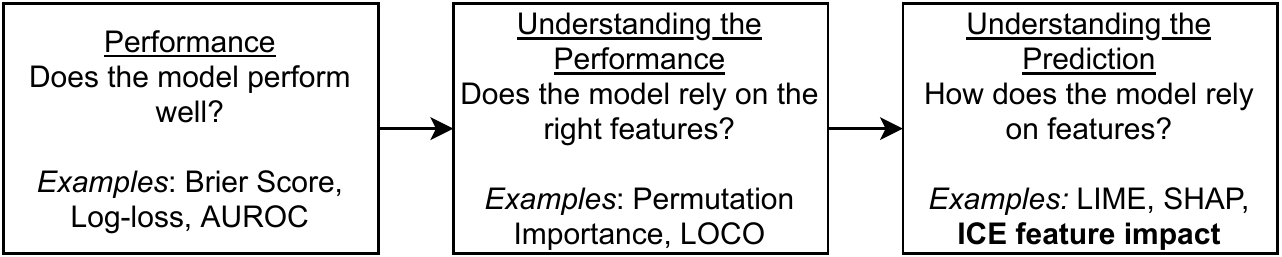}
    \caption{Three stages in model trust}
    \label{fig:motivation}
\end{figure*}

As machine learning (ML) systems become more ubiquitous in human decision making,  transparency and interpretability have grown significantly in importance \citep{Varshney16}. Some models may not require user trust due to a low-risk nature, e.g. movie recommendation systems. Other problems don't require top performance and safely rely on highly interpretable models that may not perform as well as black box models. However, when a problem space combines a high risk nature with demands for superior performance, earning the user's trust in the model is essential.
\par
We distinguish three phases to ``trusting'' a model: strong performance, model understanding, and prediction understanding (See Figure \ref{fig:motivation}). To distinguish a feature's contribution to model performance from its contribution to model predictions, we call the former ``feature importance'' and the latter ``feature impact'' \cite{parr2020}.
\par
There exist several visual methods to display feature impact, the relationship between features and predictions, most notably partial dependence plots (PDPs) \cite{friedman2001greedy} and jndividual conditional expectation (ICE) plots \cite{goldstein2014peeking}. PDPs aggregate the effects of a feature while ICE plots disaggregate divergent effects by plotting individual observations. 
\par
Visual tools are highly intuitive and can convey a lot of information in a single plot. However, they have some weaknesses as well. Firstly, visual interpretation is imprecise which makes comparison between features difficult. Secondly, ICE plots in particular can only plot a subset of the observations in the dataset to avoid overcrowding, which can hide outlier observations or overfit extrapolations from view. Thirdly, the cost of visual inspection does not scale well to the number of features--visually inspecting the plots for millions of features, for example, is infeasible.
\par
In this paper, we address these issues and extend ICE plots by extracting feature impact metrics from them (``ICE feature impact''). ICE feature impact is model- and performance-agnostic, meaning it measures the impact of each feature on the prediction only, without regarding the accuracy of that prediction. ICE feature impact also addresses the issues with the visual approach discussed above: it is a precise metric, allowing comparisons between different ICE plots; it takes into account every observation, including outliers, instead of only a subset; and it can be ranked to prioritize inspection of ICE plots to only the most impactful features, allowing the usefulness of ICE plots to scale with the number of features.
\par
We also introduce an in-distribution version of feature impact with a hyperparameter to reduce the influence of out-of-distribution points, and we supplement ICE feature impact with measures of heterogeneity and non-linearity to add depth. Together, these metrics provide a quantitative perspective for understanding feature impact complementary to the qualitative nature of inspecting ICE plots.

\section{Related Work}
\label{sec:related_work}

    First introduced by \citet{friedman2001greedy}, partial dependence plots (PDPs) are a model and performance agnostic method of illustrating the relationships between one or more input variables and the predictions of a black-box model. PDPs estimate the partial dependency by marginalizing over all other features -- essentially permuting the at-issue features to specific values across the observed range and then averaging the resulting predictions across training observations.
    \par
    Individual Conditional Expectation (ICE) \cite{goldstein2014peeking} plots disaggregate the average feature impact curve of PDPs into its component, individual observation-curves. This allows ICE plots to capture heterogeneous relationships that PDPs otherwise miss. We further discuss ICE plots and provide a specific methodology in Section \ref{sec:ice_replication}. Accumulated Local Effects \cite{ale} extend PDPs by restricting the permutation of at-issue features within a certain interval as opposed to allowing them to permute from the minimum and maximum possible values as PDPs and ICE plots do. This addresses a weakness in PDPs and ICE plots that permuting the feature value can lead to unrealistic observations when features are correlated and motivates in-distribution ICE feature impact.
    \par
    \citet{parr2020} distinguishes the idea of ``feature impact" from standard feature importance metrics as follows: while feature importance metrics measure how important a feature is to the model's performance, feature impact metrics measure how variations in feature values impact the prediction, irrespective of performance. 
    \par
    LIME \cite{lime} uses an interpretable surrogate model to approximate the feature impact on a local scale around the prediction. \citet{parr2020} proposes a non-parametric feature impact methodology that does not interrogate a fitted model. Instead, they extend the concept of PDPs by calculating the empirical partial dependence of the prediction on the at-issue feature based on the data and then approximating the area under the resulting partial dependence curve with a Riemann's Sum.
    \par
    Shapley values \cite{Sha53} detail how to fairly determine the total contribution of each feature to the overall prediction--making it a feature impact metric--by taking into account both a feature's individual contribution and collaborative contribution together with all possible subsets of features. Shapley values themselves are highly computationally expensive to calculate precisely, though they can be approximated with a Monte Carlo approach \cite{shapley_estimation}, Kernel SHAP \cite{kernel_shap}, or Tree SHAP \cite{tree_shap}. Tree SHAP differs from other approaches as it relies solely on the training data without interventionist means like permuting the value of features.
    
\section{Methodology}
\label{sec:ice}

An implementation of ICE feature impact as described below is available in Github.\footnote{\url{https://github.com/mixerupper/mltools-fi_cate}}.

\subsection{ICE Plot Replication}
\label{sec:ice_replication}

\noindent We establish terminology and notation for the remainder of the paper by detailing the ICE replication methodology we use. To replicate ICE plots, we create ``phantom observations" from each ``real observation" where all not ``at-issue features(s)" are constant, but we permute the ``at-issue feature(s)". We then use the phantom observations to interrogate the model.
\par
The exact algorithm is as follows: for at-issue feature(s) $\mathbf{x}_S$, fitted model $\hat{f}$, and feature matrix $\mathbf{X} \in \mathbb{R}^{N \times p}$, let there be $n_{\mathbf{x}_S}$ unique values of $\mathbf{x}_S$ found in the data.
\begin{enumerate}
    \item For each observation $x^{(i)}$, create $\mathbf{n_{x_S}}$ observations with all features the same as in $x^{(i)}$, except for $\mathbf{x_S}$. Replace $\mathbf{x_S}$ with the $n_{\mathbf{x_S}}$ unique values of feature $p$ found above. This results in $n_{\mathbf{x_S}}$ new observations for each $x^{(i)}$.
    \item We call the resulting observations ``phantom observations'', denoted $x^{(i)}[k]$ which is the $k$th phantom observation for $x^{(i)}$ with $k = 1,\ldots, n_{\mathbf{x_S}}$. For each observation $x^{(i)}$, one of its phantom observations is exactly identical to $x^{(i)}$, and the others are identical except for a permuted $\mathbf{x_S}$. Combine all $n\cdot n_{\mathbf{x_S}}$ phantom observations into a new feature matrix. 
    \item Use fitted model $\hat{f}$ to predict $\hat{y}$ for all phantom observations.
    \item For each original observation, plot a line composed of the corresponding phantom points with the at-issue feature on the x-axis and $\hat{y}$ on the y-axis. This results in $n$ lines, with each line composed of $n_{\mathbf{x_S}}$ phantom points.
\end{enumerate}
Additionally, if $n$ is large, we sample uniformly from each quantile of $\mathbf{x_S}$ if $\mathbf{x_S}$ is continuous and each value of $\mathbf{x_S}$ if $\mathbf{x_S}$ is categorical to capture the whole distribution.

    \subsection{ICE Feature Impact}
    \label{sec:ice_fi}
     While ICE plots allow visual inspection of feature impact, it does not output any quantitative metrics for comparability. We elicit a numeric feature impact metric from ICE plots in the form of ICE feature impact.
     \par
     For the sequence of points that make up each observation-curve, we calculate the absolute change in prediction divided by the change in feature ($|\frac{\mathrm{d}y}{\mathrm{d}x}|$) for each consecutive point. This uses rise over run to quantify the impact of the feature on the prediction value. Then, ICE feature impact is the mean of all the $|\frac{\mathrm{d}y}{\mathrm{d}x}|$ terms over all phantom points that make up an observation and all observations. To account for features of different scales, we multiply by the standard deviation of that feature. We will see that ICE feature impact has an analogous interpretation to coefficients in a linear model. 
    \par
    The exact algorithm is as follows: for feature $\mathbf{x}_S$, let $\sigma_{\mathbf{x}_S}$ denote the standard deviation of $\mathbf{x}_S$, let $n$ be the number of observations, $n_{\mathbf{x}_S}$ be the number of unique values of $\mathbf{x}_S$, $x^{(i)}$ be the ith observation, $x^{(i)}[k]$ be the $k$th phantom observation corresponding to $x^{(i)}$, $\mathbf{x}_S^{(i)}$ be the value of $\mathbf{x}_S$ in observation $x^{(i)}$, $\mathbf{x}_S^{(i)}[k]$ be the value of $\mathbf{x}_S$ in the $k$th phantom observation corresponding to $x^{(i)}$, and $\hat{y}$ be the fitted model. Thus, the {\bf ICE feature impact} is:
    
    \begin{equation} \label{eq:feature_impact}
        \begin{split}
            \textbf{FI}(\mathbf{x}_S) &= \frac{\sigma_{\mathbf{x}_S}}{n\cdot (n_{\mathbf{x}_S} - 1)}\sum_{i=1}^{n}\sum_{k = 2}^{n_{\mathbf{x}_S}}\bigg|\frac{d\hat{y}(x^{(i)}[k])}{dx_S^{(i)}[k]}\bigg|\\
            &\approx \frac{\sigma_{\mathbf{x}_S}}{n\cdot (n_{\mathbf{x}_S} - 1)}\sum_{i=1}^{n}\sum_{k = 2}^{n_{\mathbf{x}_S}}\bigg|\frac{\hat{y}(x^{(i)}[k]) - \hat{y}(x^{(i)}[k-1])}{x_S^{(i)}[k] - x_S^{(i)}[k-1]}\bigg|\\
        \end{split}
    \end{equation}
    
    \noindent The ICE feature impact of $\mathbf{x}_S$ can be interpreted as the absolute change in the predicted value of $\hat{y}$ for each one-unit change in $\mathbf{x}_S$ if $\mathbf{x}_S$ was standardized to a standard deviation of $1$ and all other features remained constant. Note that ICE feature impact gives the magnitude of impact, not the direction. Average direction of feature impact can be determined by comparing the ICE feature impact with the value of Equation \ref{eq:feature_impact} without an absolute value on the inner summation term.
    
    \subsection{In-Distribution ICE Feature Impact}
    
    One of the drawbacks of ICE feature impact as introduced in \Cref{sec:ice_fi} is that it weights evenly across all phantom points, no matter their likelihood of occurrence in the true feature distribution. This may be concerning if features are highly correlated, and permuting the at-issue feature $\mathbf{x_S}$ takes us out of the feature distribution, e.g., taking the health data from a 9 year old and changing the age to 70 while leaving the other features untouched would give us a phantom observation that has a low likelihood of occurring in reality.
    \par
    This is a missing data problem with the missing value being the likelihood of the observation. The likelihood is $1$ for all true observations and missing for all phantom observations. Let us denote the likelihood of phantom observation $x^{(i)}[k]$ for at-issue feature $\mathbf{x}_S$ with $L_{\mathbf{x}_S}(x^{(i)}[k])$. Then, given this likelihood, the in-distribution ICE feature impact of $\mathbf{x}_S$ is:
    
    \begin{equation} \label{eq:in_dist_feature_impact}
        \begin{split}
            \textbf{IDFI}(\mathbf{x}_S) &\approx \frac{\sigma_{\mathbf{x}_S}}{\sum_{i=1}^{n}\sum_{k = 2}^{n_{\mathbf{x}_S}}L_{\mathbf{x}_S}}\sum_{i=1}^{n}\sum_{k = 2}^{n_{\mathbf{x}_S}}L_{\mathbf{x}_S}(x^{(i)}[k])\bigg|\frac{\hat{y}(x^{(i)}[k]) - \hat{y}(x^{(i)}[k-1])}{\mathbf{x}_S^{(i)}[k] - x^{(i)}_S[k-1]}\bigg|\\
        \end{split}
    \end{equation}
    \\
    To estimate $L_{\mathbf{x}_S}(x^{(i)}[k])$, we model likelihood as exponentially decaying with respect to the absolute distance of the at-issue feature's permutation divided by the feature's standard deviation:
    \begin{equation} \label{eq:in_dist_likelihood}
        \begin{split}
            L_{\mathbf{x}_S}(x^{(i)}[k]) = \lambda ^{\frac{|\mathbf{x}_S^{(i)}[k] - \mathbf{x}_S^{(i)}|}{\sigma_{\mathbf{x}_S}}}
        \end{split}
    \end{equation}
    where $0 < \lambda \leq 1$ is a hyperparameter that measures how quickly the weight decays as the phantom feature value differs from the real feature value. Note that $\lambda = 1$ gets us back to ICE feature impact without out-of-distribution considerations. 
    \par
    We can estimate $\sigma_{\mathbf{x}_S}$ as the sample standard deviation of $\mathbf{x}_S$ in the data or as an arbitrarily sophisticated estimate of the standard deviation for the at-issue feature based on the value of all other features for the observation. For example, \cite{fast_accurate_simple_models} proposes estimating the conditional distribution of a feature based on all other features using a pseudo-maximum likelihood problem estimated via a single self-attention architecture.
    \par
    The in-distribution ICE feature impact weights phantom observations closer to the real observation more heavily when measuring feature impact, giving us a perspective on feature impact that is more ``true to the data'' \cite{true_to_data_or_model}.

\section{Real Data}
\label{sec:experiments}

To examine ICE feature impact, we use UC Irvine's cervical cancer risk factors dataset.\footnote{\href{https://archive.ics.uci.edu/ml/datasets/Cervical+cancer+\%28Risk+Factors\%29}{Cervical Cancer (Risk Factors) Data Set} contains a detailed description of the dataset.} The dataset contains medical information for 858 patients from \textit{Hospital Universitario de Caracas}. There are 32 features including age, number of pregnancies, and use of IUD. The target variable is \texttt{Biopsy}, which is binary.

    \subsection{Complementary to Feature Importance}
    First, we show that ICE feature impact presents an additional dimension to understanding models beyond feature importance.
    \par
    We train a random forest classifier \cite{breiman2001random} on the dataset.\footnote{We use the \texttt{sklearn} package with parameters of 500 trees, a random state seed of 20, and the default values for the remaining parameters. As this exercise is about model interpretability, we did not tune the model to improve performance.} We then calculate the following metrics for each feature: ICE feature impact, Tree SHAP \cite{tree_shap}, Random Forest feature importance \cite{breiman_2002_manual}, and permutation feature importance \cite{breiman2001random} and normalize them to be positive and sum to 100. We take the correlation between ICE feature impact and alternative metrics and find that the correlation is low (See Table \ref{tab:ice_corr}). This indicates that ICE feature impact differs substantially from alternatives instead of fulfilling the same function.
    
    \begin{table}
    \centering
        \begin{tabular}{l>{\centering\arraybackslash}m{2cm}}
            \hline
            Metric & Correlation w/ ICE FI \\
            \hline
            In-Distribution ICE FI ($\lambda = 0.75$) &  0.99 \\
            Random Forest Feature Importance & 0.36 \\
            Permutation Feature Importance & 0.35 \\
            Tree SHAP Values & 0.17\\
            \hline
        \end{tabular}
        \caption{Pearson correlation of feature importance and impact metrics with ICE feature impact. All metrics were first normalized to sum to 100. Tree SHAP values were additionally first made positive to remove direction before normalizing to sum to 100.}
        \label{tab:ice_corr}
    \end{table}
    
    Table \ref{tab:fi_diffs} shows the features with the two most positive differences and the features with the two most negative differences between their random forest feature importance and ICE feature impact values.\footnote{See \ref{appendix:fi_table} for the full feature impact table, \ref{appendix:ice_plots} for the ICE plots for all features, and \ref{appendix:c-ice_plots} for the centered ICE plots (c-ICE) \cite{goldstein2014peeking} for all features.} While \texttt{Age} and \texttt{Number of Sexual Partners} are highly predictive features and are helpful in reducing impurity of classification, they do not have a strong impact on the model's predictions itself. On the opposite end of the spectrum, \texttt{STDs:molluscum contagiosum} and \texttt{STDs:pelvic inflammatory disease} have highly imbalanced feature distributions with the majority of values equal to 0 and therefore are not as helpful for reducing impurity. However, when these factors are present -- specifically, when the value is missing and the mean is imputed -- they contribute strongly to the model prediction, explaining the higher feature impact.
    
\begin{table}
\centering
\begin{tabular}{m{4.5cm}>{\centering\arraybackslash}m{1.5cm}>{\centering\arraybackslash}m{1.5cm}>{\centering\arraybackslash}m{1.5cm}}
\toprule
                          Feature &  ICE FI &  Native Feature Importance &  Difference \\
\midrule
       STDs:molluscum contagiosum &     9.8 &                        0.1 &         9.6 \\
 STDs:pelvic inflammatory disease &     8.9 &                        0.1 &         8.8 \\
        Number of sexual partners &     1.0 &                        9.9 &        -8.9 \\
                              Age &     3.4 &                       17.8 &       -14.3 \\
\bottomrule
\end{tabular}
\caption{Feature impact table for features in cervical cancer dataset with two largest and most negative difference between Random Forest feature importance and ICE Feature Impact.}
\label{tab:fi_diffs}
\end{table}
    
     \subsection{Interpretability: Analogous to Linear Regression Coefficients}
    In the base case of analyzing a linear regression model, ICE feature impact values are exactly the absolute value of the linear regression coefficients. We also calculated ICE feature impact for the pseudo-linear models of Logistic Regression and linear SVMs. Table \ref{tab:linear_vs_ice} shows that the resulting model coefficients are strongly correlated with the corresponding ICE feature impact values.
    
    \begin{table}
    \centering
        \begin{tabular}{lcc}
                &\multicolumn{2}{c}{\textbf{ICE Feature Impact}} \\
                \hline
                Model & Base & In-Dist \\
                \hline
                Linear Regression & 1 & 1 \\
                Logistic & 0.73 & 0.8\\
                SVM & 0.9 & 0.98\\
                \hline
        \end{tabular}
        \caption{Pearson correlation of ICE feature impact values with absolute value of coefficients of linear and pseudo-linear models.}
        \label{tab:linear_vs_ice}
    \end{table}
    
    These results show that ICE feature impact can be interpreted analogously to linear regression coefficients with features standardized to a unit standard deviation.
    
    \subsection{Quantifying Heterogeneity and Non-Linearity}
    In linear models, knowing feature impact means knowing exactly where predictions come from. In non-linear models, however, the relationship between features and the model prediction can be more complex: in particular, the relationship can be heterogeneous -- different across observations -- or non-linear -- different across the feature's support. We propose measures of heterogeneity and non-linearity to allow the practitioner a more nuanced understanding of ICE feature impact.
    \par
    Let heterogeneity be the degree to which the pattern of ICE curves varies across observations, i.e. the feature impact is heterogeneous when its impact is higher on some observations and lower on others. Then, following the notation described in Section \ref{sec:ice_replication}, the heterogeneity of feature $\mathbf{x}_S$ is:
    \begin{equation} \label{eq:heterogeneity}
        \begin{split}
            \textbf{HE}(\mathbf{x}_S) &= \frac{\sigma_{\mathbf{x}_S}}{n_{\mathbf{x}_S}}\sum_{k=1}^{n_{\mathbf{x}_S}}SD_{i \in \{1,\ldots,n\}}\bigg(\frac{\hat{y}(x^{(i)}[k]) - \hat{y}(x^{(i)}[k-1])}{x_S^{(i)}[k] - x_S^{(i)}[k-1]}\bigg)\\
        \end{split}
    \end{equation}
    
    \noindent where the standard deviation is taken for fixed $k$ across all real observations. The lower the heterogeneity metric, the more similar the shape of observation-curves are at each point. For linear regressions and additive models like GAM \cite{gam}, the heterogeneity metric is zero since the effect of a feature on the prediction is the same across all observations.
    \par
    Let non-linearity be the degree to which features have a non-linear relationship with the model's predictions, i.e. how much the effect of a feature varies across the support for a given observation. For features with low non-linearity, the corresponding ICE feature impact can be interpreted as close to a linear regression coefficient, even if the underlying model is non-linear. We quantify non-linearity as follows: 
    \begin{equation} \label{eq:non-linearity}
        \begin{split}
            \textbf{NL}(\mathbf{x}_S) &= \frac{\sigma_{\mathbf{x}_S}}{n}\sum_{i=1}^{n}SD_{k \in \{1,\ldots, n_{\mathbf{x}_S}\}}\bigg(\frac{\hat{y}(x^{(i)}[k]) - \hat{y}(x^{(i)}[k-1])}{x_S^{(i)}[k] - x_S^{(i)}[k-1]}\bigg)\\
        \end{split}
    \end{equation}
    
    \noindent where the standard deviation is taken for fixed $i$ across all corresponding phantom observations. For linear regressions, the non-linearity is equal to 0 as desired since the effect of a feature is constant across the feature's support.
    \par
    Table \ref{tab:het_and_non-lin} shows the heterogeneity and non-linearity of the features listed in Table \ref{tab:fi_diffs}.\footnote{See \ref{appendix:deriv_fi_table} for heterogeneity and non-linearity for all features.} Note that the features with the largest positive differences between feature impact and feature importance have higher heterogeneity but similar non-linearity compared to the features with the largest negative differences. This is because ICE feature impact captures heterogeneity through taking the absolute value of the feature impact $\frac{dy}{dx}$ units but does not discriminate between non-linear or linear relationships.
    
\begin{table}
\centering
\begin{tabular}{m{4.5cm}>{\centering\arraybackslash}m{2cm}>{\centering\arraybackslash}m{2cm}>{\centering\arraybackslash}m{2cm}}
\toprule
                          Feature &  Feature Impact &  Heterogeneity &  Non-Linearity \\
\midrule
       STDs:molluscum contagiosum &             9.8 &           0.27 &           0.19 \\
 STDs:pelvic inflammatory disease &             8.9 &           0.23 &           0.17 \\
        Number of sexual partners &             1.0 &           0.05 &           0.04 \\
                              Age &             3.4 &           0.11 &           0.18 \\
\bottomrule
\end{tabular}
\caption{ICE feature impact, heterogeneity, and non-linearity for features in cervical cancer dataset with the two most positive and most negative differences between Random Forest feature importance and ICE Feature Impact.}
\label{tab:het_and_non-lin}
\end{table}

\section{Discussion}
\label{sec:discussion}
Building upon efforts to interpret machine learning models, we extend ICE plots by drawing out ICE feature impact, a measure of the relationship between features and model predictions. ICE feature impact is uncorrelated with alternative feature importance metrics, highlighting features that are impactful to predictions but do not contribute as strongly to model performance. It has a highly interpretable form and is analogous to linear regression coefficients. 
\par
We also propose in-distribution ICE feature impact to downweight out-of-distribution observations and the heterogeneity and non-linearity measures that add dimensionality to our characterization of ICE feature impact.
\par
Altogether, ICE feature impact provides a different perspective from traditional feature importance methods, complements ICE plots, and serves as an alternative to SHAP values in understanding where a model's predictions come from.

\newpage

% \bibliographystyle{splncs04}
% \bibliography{mybibliography}

\bibliographystyle{splncs04}
\bibliography{ref}

\appendix
\renewcommand{\thesection}{\appendixname~\Alph{section}}

\newpage
\section{ICE Plots for Cervical Cancer Data}
\label{appendix:ice_plots}

\begin{figure}[!ht]
    \centering
    \includegraphics[trim = 0in 19in 0in 0in, clip,width=\textwidth,height=0.78\textheight,keepaspectratio]{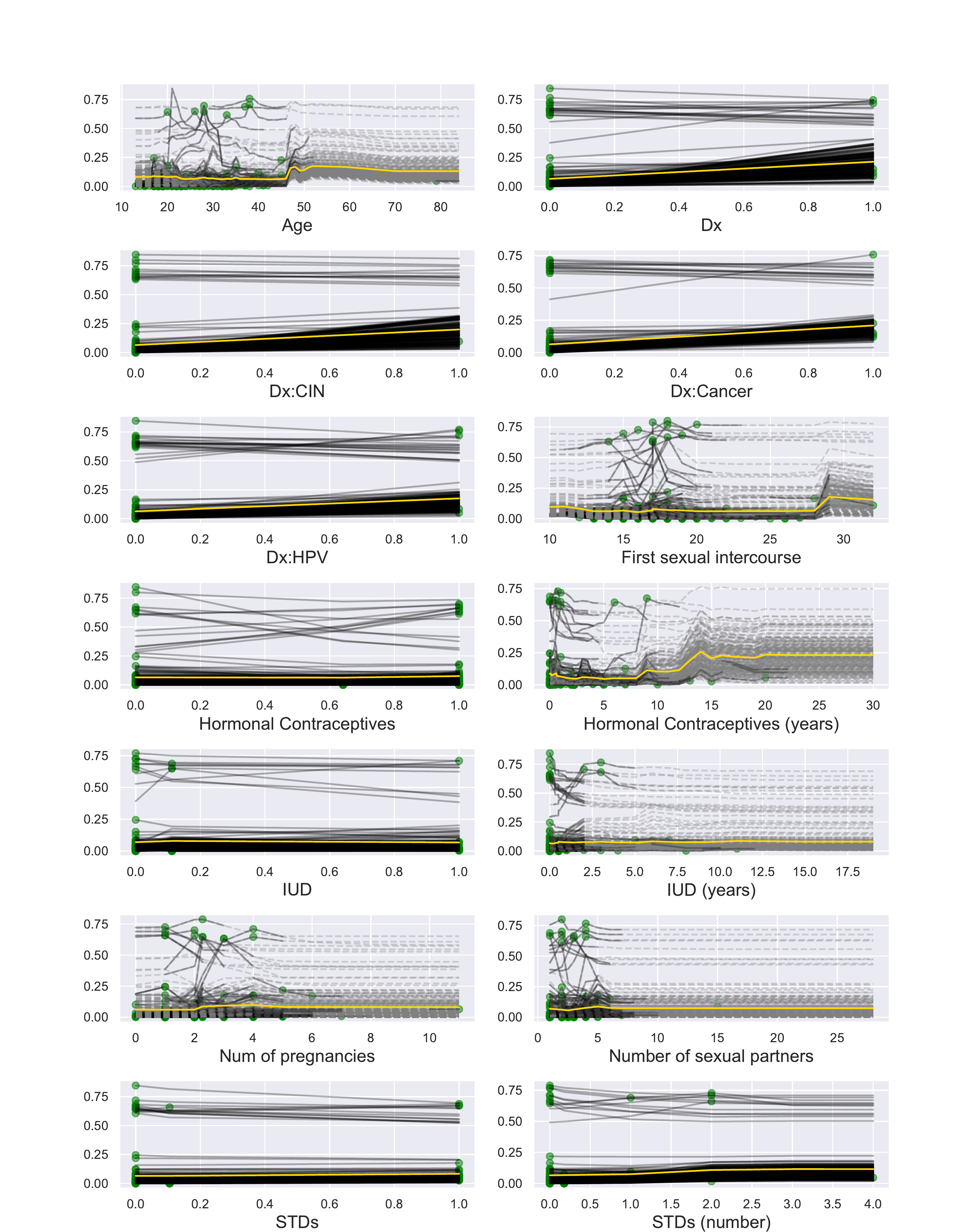}
    \caption{ICE plots for all features in cervical cancer dataset, following the methodology described in Section \ref{sec:ice_replication}. Each green dot represents a different observation, and the corresponding line shows how varying the observation's at-issue feature value affects the model's prediction. Observation-lines are solid within $\frac{1}{2}$ a standard deviation (of the at-issue feature) and dotted outside that range.}
    \label{fig:cancer-all-ice}
\end{figure}
\newpage
\begin{center}
\includegraphics[trim = 0in 0in 0in 14.8in, clip,width=\textwidth,height=\textheight,keepaspectratio]{cervical-cancer/all-ice-plots.pdf}
\end{center}

\newpage
\section{c-ICE Plots for Cervical Cancer Data}
\label{appendix:c-ice_plots}

\begin{figure}[!ht]
    \centering
    \includegraphics[trim = 0in 17in 0in 0in, clip,width=\textwidth,height=0.83\textheight,keepaspectratio]{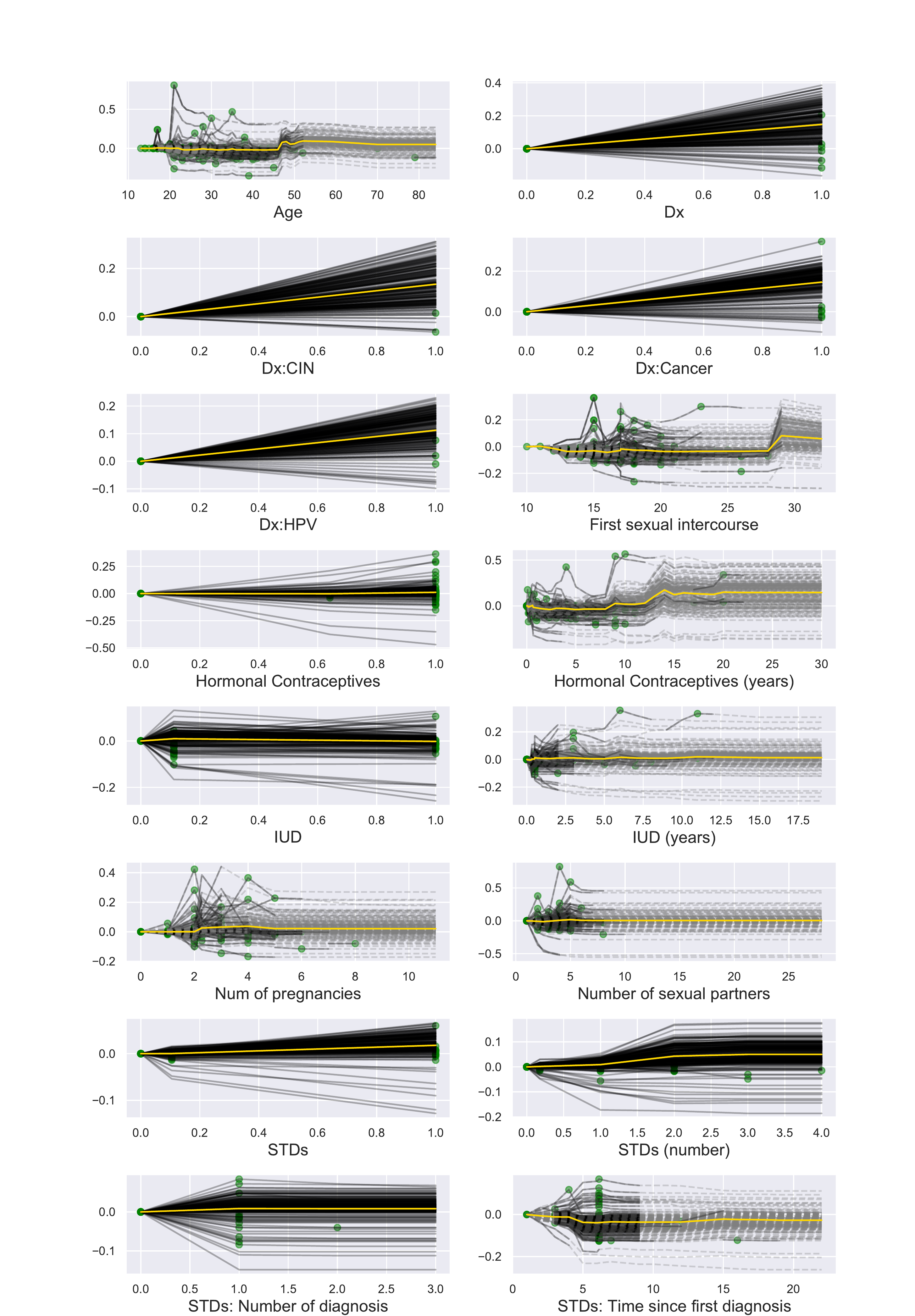}
    \caption{Centered ICE (c-ICE) plots \cite{goldstein2014peeking} for all features in cervical cancer dataset. c-ICE plots are equivalent to ICE plots but with the starting $\hat{y}$ value centered to zero such that the lines represent the change in $\hat{y}$ instead of its value.}
    \label{fig:cancer-all-c-ice}
\end{figure}
\newpage
\begin{center}
    \includegraphics[trim = 0in 0in 0in 16.9in, clip,width=\textwidth,height=0.83\textheight,keepaspectratio]{cervical-cancer/all-c-ice-plots.pdf}
\end{center}

\newpage
\section{Feature Impact Table for Cervical Cancer Data}
\label{appendix:fi_table}
\FloatBarrier
\begin{table}[!hbt]
\centering
\label{app:all_fi}
\begin{tabular}{m{4.5cm}>{\centering\arraybackslash}m{1.5cm}>{\centering\arraybackslash}m{1.5cm}>{\centering\arraybackslash}m{1.5cm}>{\centering\arraybackslash}m{1.5cm}>{\centering\arraybackslash}m{1.5cm}}
\toprule
                        Feature &   ICE &  ICE ID ($\lambda = 0.75$) &  Random Forest &  Tree SHAP &  Permutation \\
\midrule
                            Age &   3.4 &                        2.6 &           17.8 &       11.3 &         13.3 \\
      Number of sexual partners &   1.0 &                        1.1 &            9.9 &        6.3 &          9.9 \\
       First sexual intercourse &  16.9 &                       16.9 &           12.3 &        7.2 &         12.1 \\
             Num of pregnancies &   1.3 &                        1.4 &           10.0 &        0.0 &         13.1 \\
                         Smokes &   1.3 &                        1.1 &            1.4 &        0.2 &          1.9 \\
                 Smokes (years) &   1.8 &                        1.8 &            3.9 &        6.8 &          4.6 \\
            Smokes (packs/year) &   7.0 &                        6.6 &            3.7 &        1.9 &          4.4 \\
        Hormonal Contraceptives &   0.8 &                        0.6 &            2.9 &       10.8 &          5.1 \\
 Hormonal Contraceptives (ye... &   9.0 &                        7.8 &           15.6 &        7.8 &         15.9 \\
                            IUD &   2.2 &                        2.0 &            2.2 &        3.4 &          3.0 \\
                    IUD (years) &   3.7 &                        4.0 &            3.6 &        4.6 &          4.6 \\
                           STDs &   0.6 &                        0.4 &            0.5 &        0.7 &          0.1 \\
                  STDs (number) &   0.6 &                        0.6 &            1.1 &        1.2 &          1.0 \\
            STDs:condylomatosis &   1.2 &                        1.0 &            0.6 &        1.8 &          0.5 \\
   STDs:cervical condylomatosis &   0.0 &                        0.0 &            0.0 &        0.0 &          0.0 \\
    STDs:vaginal condylomatosis &   3.8 &                        4.1 &            0.3 &        0.8 &          0.0 \\
 STDs:vulvo-perineal condylo... &   1.1 &                        0.9 &            0.6 &        1.5 &          0.5 \\
                  STDs:syphilis &   2.1 &                        2.2 &            0.4 &        1.1 &          0.0 \\
 STDs:pelvic inflammatory di... &   8.9 &                        9.9 &            0.1 &        1.3 &          0.0 \\
            STDs:genital herpes &   6.8 &                        7.4 &            0.9 &        2.4 &          0.5 \\
     STDs:molluscum contagiosum &   9.8 &                       10.8 &            0.1 &        1.0 &          0.0 \\
                      STDs:AIDS &   0.0 &                        0.0 &            0.0 &        0.0 &          0.0 \\
                       STDs:HIV &   1.2 &                        1.2 &            1.1 &        1.8 &          1.9 \\
               STDs:Hepatitis B &   5.7 &                        6.3 &            0.2 &        1.5 &          0.0 \\
                       STDs:HPV &   5.4 &                        5.9 &            0.1 &        1.0 &          0.0 \\
      STDs: Number of diagnosis &   0.2 &                        0.2 &            0.7 &        4.7 &          0.0 \\
 STDs: Time since first diag... &   0.3 &                        0.3 &            2.0 &        1.2 &          1.5 \\
 STDs: Time since last diagn... &   0.3 &                        0.3 &            1.7 &        2.7 &          0.0 \\
                      Dx:Cancer &   1.0 &                        0.7 &            1.7 &        3.5 &          2.1 \\
                         Dx:CIN &   0.6 &                        0.5 &            1.2 &        3.9 &          1.1 \\
                         Dx:HPV &   0.8 &                        0.6 &            1.6 &        4.0 &          1.6 \\
                             Dx &   1.2 &                        0.9 &            1.7 &        3.7 &          1.1 \\
\bottomrule
\end{tabular}
\caption{Feature impact table for all features in cervical cancer dataset. All feature impact/importance metrics have been made positive and normalized to sum to 100. The ordering of features is as ordered in the original dataset.}
\end{table}

\newpage
\section{Heterogeneity and Non-Linearity of ICE Feature Impact for Cervical Cancer Data}
\label{appendix:deriv_fi_table}
\FloatBarrier

\begin{table}[!ht]
\centering
\label{app:fi_deriv}
\begin{tabular}{m{4.5cm}>{\centering\arraybackslash}m{2cm}>{\centering\arraybackslash}m{2cm}>{\centering\arraybackslash}m{2cm}}
\toprule
                        Feature &  ICE Feature Impact &  ICE Heterogeneity &  ICE Non-Linearity \\
\midrule
                            Age &                0.08 &               0.11 &               0.18 \\
      Number of sexual partners &                0.02 &               0.05 &               0.04 \\
       First sexual intercourse &                0.38 &               0.59 &               1.63 \\
             Num of pregnancies &                0.03 &               0.05 &               0.07 \\
                         Smokes &                0.03 &               0.04 &               0.02 \\
                 Smokes (years) &                0.04 &               0.06 &               0.12 \\
            Smokes (packs/year) &                0.16 &               0.32 &               0.60 \\
        Hormonal Contraceptives &                0.02 &               0.04 &               0.01 \\
 Hormonal Contraceptives (ye... &                0.20 &               0.34 &               0.47 \\
                            IUD &                0.05 &               0.07 &               0.04 \\
                    IUD (years) &                0.08 &               0.12 &               0.28 \\
                           STDs &                0.01 &               0.02 &               0.01 \\
                  STDs (number) &                0.01 &               0.02 &               0.01 \\
            STDs:condylomatosis &                0.03 &               0.03 &               0.01 \\
   STDs:cervical condylomatosis &                0.00 &               0.00 &               0.00 \\
    STDs:vaginal condylomatosis &                0.08 &               0.11 &               0.07 \\
 STDs:vulvo-perineal condylo... &                0.02 &               0.03 &               0.01 \\
                  STDs:syphilis &                0.05 &               0.07 &               0.04 \\
 STDs:pelvic inflammatory di... &                0.20 &               0.23 &               0.17 \\
            STDs:genital herpes &                0.15 &               0.18 &               0.13 \\
     STDs:molluscum contagiosum &                0.22 &               0.27 &               0.19 \\
                      STDs:AIDS &                0.00 &               0.00 &               0.00 \\
                       STDs:HIV &                0.03 &               0.03 &               0.02 \\
               STDs:Hepatitis B &                0.13 &               0.17 &               0.11 \\
                       STDs:HPV &                0.12 &               0.14 &               0.10 \\
      STDs: Number of diagnosis &                0.00 &               0.00 &               0.00 \\
 STDs: Time since first diag... &                0.01 &               0.01 &               0.02 \\
 STDs: Time since last diagn... &                0.01 &               0.01 &               0.02 \\
                      Dx:Cancer &                0.02 &               0.01 &               0.00 \\
                         Dx:CIN &                0.01 &               0.01 &               0.00 \\
                         Dx:HPV &                0.02 &               0.01 &               0.00 \\
                             Dx &                0.03 &               0.02 &               0.00 \\
\bottomrule
\end{tabular}
\caption{Heterogeneity and non-linearity dimensions for all features in cervical cancer dataset. The raw ICE feature impact is presented without normalization to sum to 100.}
\end{table}

\end{document}